\documentclass[]{article}
\pdfoutput=1

\usepackage{PRIMEarxiv}

\usepackage[utf8]{inputenc} 
\usepackage[T1]{fontenc}    
\usepackage{hyperref}       
\usepackage{url}            
\usepackage{booktabs}       
\usepackage{amsfonts}       
\usepackage{nicefrac}       
\usepackage{microtype}      
\usepackage{lipsum}
\usepackage{fancyhdr}       
\usepackage{graphicx}       
\usepackage{multicol}
\usepackage{placeins}

\usepackage[
backend=biber,
style=ieee,
citestyle=numeric-comp
]{biblatex}
\AtEveryBibitem{\clearfield{address}}
\AtEveryBibitem{\clearlist{address}}

\AtEveryBibitem{\clearfield{place}}
\AtEveryBibitem{\clearlist{place}}

\AtEveryBibitem{\clearfield{issn}}
\AtEveryBibitem{\clearlist{issn}}

\AtEveryBibitem{\clearfield{series}}
\AtEveryBibitem{\clearlist{series}}

\AtEveryBibitem{\clearfield{month}}
\AtEveryBibitem{\clearlist{month}}

\AtEveryBibitem{\clearfield{isbn}}
\AtEveryBibitem{\clearlist{isbn}}

\AtEveryBibitem{\clearfield{language}}
\AtEveryBibitem{\clearlist{language}}

\AtEveryBibitem{\clearfield{doi}}
\AtEveryBibitem{\clearlist{doi}}

\title{Large Language Models Can Infer Psychological Dispositions of Social Media Users}

\author{
  Heinrich Peters\\
  Columbia University\\
  New York\\
  \texttt{\ hp2500@columbia.edu}\\
   \And
  Sandra Matz \\
  Columbia University\\
  New York\\
  \texttt{\ sm4409@columbia.edu }\\
}

\begin{document}

\maketitle

\begin{abstract}
Large Language Models (LLMs) demonstrate increasingly human-like abilities across a wide variety of tasks. In this paper, we investigate whether LLMs like ChatGPT can accurately infer the psychological dispositions of social media users and whether their ability to do so varies across socio-demographic groups. Specifically, we test whether GPT-3.5 and GPT-4 can derive the Big Five personality traits from users' Facebook status updates in a zero-shot learning scenario. Our results show an average correlation of r = .29 (range = [.22, .33]) between LLM-inferred and self-reported trait scores – a level of accuracy that is similar to that of supervised machine learning models specifically trained to infer personality. Our findings also highlight heterogeneity in the accuracy of personality inferences across different age groups and gender categories: predictions were found to be more accurate for women and younger individuals on several traits, suggesting a potential bias stemming from the underlying training data or differences in online self-expression. The ability of LLMs to infer psychological dispositions from user-generated text has the potential to democratize access to cheap and scalable psychometric assessments for both researchers and practitioners. On the one hand, this democratization might facilitate large-scale research of high ecological validity and spark innovation in personalized services. On the other hand, it also raises ethical concerns regarding user privacy and self-determination, highlighting the need for stringent ethical frameworks and regulation.\end{abstract}


\section{Introduction}
Large language models (LLMs) and other transformer-based neural networks have revolutionized text analysis in research and practice. Models such as OpenAI's GPT-4 \cite{openai_gpt-4_2023} or Anthropic’s Claude \cite{anthropic_model_2023}, for example, have shown a remarkable ability to represent, comprehend, and generate human-like text. Compared to prior NLP approaches, one of the most striking advances of LLMs is their ability to generalize their “knowledge” to novel scenarios, contexts, and tasks \cite{brown_language_2020, radford_language_2019}.

While LLMs were not explicitly designed to capture or mimic elements of human cognition and psychology, recent research suggests that – given their training on extensive corpora of human-generated language – they might have spontaneously developed the capacity to do so. For example, LLMs display properties that are similar to the cognitive abilities and processes observed in humans, including theory of mind (i.e., the ability to understand the mental states of other agents \cite{kosinski_theory_2023}), cognitive biases in decision-making \cite{hagendorff_thinking_2023} and semantic priming \cite{digutsch_overlap_2023}. Similarly, LLMs are able to effectively generate persuasive messages tailored to specific psychological dispositions (e.g., personality traits, moral values \cite{matz_potential_2024}).

Here, we examine whether LLMs possess another quality that is fundamentally human: The ability to “read” people and form first impressions about their psychological dispositions in the absence of direct or prior interaction. As research under the umbrella of zero-acquaintance studies shows, people can be remarkably accurate at judging the psychological traits of strangers simply by observing traces of their behavior under certain conditions \cite{albright_consensus_1988}. While such judgments can be influenced by stereotypes and their accuracy can vary based on the traits being assessed and the context in which judgments are made \cite{kenny_consensus_1994}, past work indicates that people are able to predict a stranger’s personality traits by observing their offices or bedrooms \cite{gosling_room_2002}, examining their music preferences \cite{rentfrow_message_2006}, or scrolling through their social media profiles \cite{back_facebook_2010}.

Existing research in computational social science shows that supervised machine learning models are able to make similar predictions. That is, given a large enough dataset including both self-reported personality traits and people’s digital footprints - such as Facebook Likes, music playlists, or browsing histories – machine learning models are able to statistically relate both inputs in a way that allows them to predict personality traits after observing a person’s digital footprints \cite{kosinski_private_2013, azucar_predicting_2018}. This is also true for various forms of text data, including social media posts \cite{schwartz_personality_2013, park_automatic_2015}, personal blogs \cite{yarkoni_personality_2010}, or short text responses collected in the context of job applications \cite{grunenberg_machine_2024}.

In this paper, we test whether LLMs have the ability to make similar psychological inferences without having been explicitly trained to do so (known as zero-shot learning \cite{brown_language_2020}). Specifically, we use Open AI’s ChatGPT (GPT-3.5 and GPT-4 \cite{openai_gpt-4_2023}) to explore whether LLMs can accurately infer the Big Five personality traits Openness, Conscientiousness, Extraversion, Agreeableness, and Neuroticism \cite{mccrae_five-factor_2008} of social media users from the content of their Facebook status updates in a zero-shot scenario. In addition, we test for biases in ChatGPT’s judgments that might arise from its foundation in equally biased human-generated data. Building on previous work highlighting inherent stereotypes in pre-trained NLP models \cite{bolukbasi_man_2016, wan_biasasker_2023}, we explore the extent to which the personality inferences made by ChatGPT are indicative of gender and age-related biases (e.g., potential biases in how the personality of men and women or older and younger people is judged).

Understanding the capabilities and limitations of LLMs with regard to inferring highly intimate psychological traits from digital footprints is critical, given their rapid adoption in both research and practice. On the one hand, easy access to the psychological profiles of individuals creates unprecedented opportunities to study individual differences at scale and customize products, services, or behavioral interventions to individuals’ unique dispositions. On the other hand, however, automated psychological inferences pose considerable ethical and legal challenges with regard to individuals’ privacy and self-determination \cite{matz_privacy_2020}. This problem is exacerbated by the fact that LLMs are also able to automatically craft persuasive messages based on users' personality profiles \cite{matz_potential_2024}. The combination of fully automated psychological assessments and personalized interactions opens the door for manipulation and misuse at scale and with little to no human oversight. We discuss our findings in light of both the opportunities for scientists and practitioners and the challenges that will require new forms of AI governance and regulation \cite{perc_social_2019, hacker_regulating_2023, chan_gpt-3_2023}.

\section{Method}
\subsection{Data and Sampling}
Our analyses are based on text data obtained from MyPersonality \cite{kosinski_facebook_2015}, a Facebook application that allowed users to take real psychometric tests - including a validated measure of the Big Five personality traits (IPIP \cite{goldberg_international_2006}) - and receive immediate feedback on their responses. Users also had the opportunity to donate their Facebook profile information - including their public profiles, Facebook Likes, and status updates – to research. For the purpose of this study, we randomly subsampled 1,000 adult users (24.2 $\pm$ 8.8 years old, 63.1\% female) who completed the full 100-item IPIP personality questionnaire and had at least 200 Facebook status updates (if they had more, we used the most recent 200). The study received IRB approval from Columbia University’s ethics review board (Protocol \#AAAU8559).

\subsection{Measures}
MyPersonality measured users’ personality traits using the International Personality Item Pool (IPIP \cite{goldberg_international_2006}), a widely established self-report questionnaire that captures the Big Five personality traits of Openness, Conscientiousness, Extraversion, Agreeableness, and Neuroticism \cite{mccrae_five-factor_2008}. We only included users who had completed the full questionnaire with all 100 items.

To obtain inferred personality traits from ChatGPT, we used the last 200 Facebook status updates generated by each user without additional preprocessing. The average length of status updates in our sample was 17.10 words (SD=15.03). Status updates were scored using the ChatGPT API with GPT-3.5 (version gpt-3.5-turbo-0301) and GPT-4 (version gpt-4-0314) \cite{openai_gpt-4_2023} as underlying models. For this purpose, the status updates were first concatenated into chunks and then fed into the GPT model, using a set of simple prompts to guide the behavior of the model. The system prompt was the default for GPT-3.5 and GPT-4, respectively: “You are a helpful assistant”. Additionally, we prompted the model to infer Big Five traits using the inference prompt: “Rate the text on the Big Five personality dimensions. Pay attention to how people's personalities might be reflected in the content they post online. Provide your response on a scale from 1 to 5 for the traits Openness, Conscientiousness, Extraversion, Agreeableness, and Neuroticism. Provide only the numbers.” We then used a simple text-parsing script to transform the responses into numerical scores. In order to avoid exceeding the GPT token limit, status update histories were processed in chunks of 20 messages, and the inferred personality scores were then averaged to derive overall scores.

To boost the reliability of the inferred personality estimates, we queried ChatGPT three times for each inference. Agreement across ratings across rating rounds was high for all traits (Openness: r$_{GPT3.5}$= .88, r$_{GPT4}$= 0.73; Conscientiousness: r$_{GPT3.5}$= .88, r$_{GPT4}$= 0.91; Extraversion: r$_{GPT3.5}$= .92, r$_{GPT4}$= 0.87; Agreeableness: r$_{GPT3.5}$= .96, r$_{GPT4}$= 0.94; Neuroticism: r$_{GPT3.5}$= .91, r$_{GPT4}$= 0.93), and all p-values were smaller than .001 with Bonferroni correction for multiple comparisons. Given the high level of agreement, we computed aggregate inferred scores by averaging scores across the three rounds of rating. We used the aggregate scores for all further analyses.

\section{Results}
\subsection{Can LLMs Infer Personality Traits From Social Media Posts?}
\begin{figure*}[t!]
    \centering
    \includegraphics[width=\textwidth]{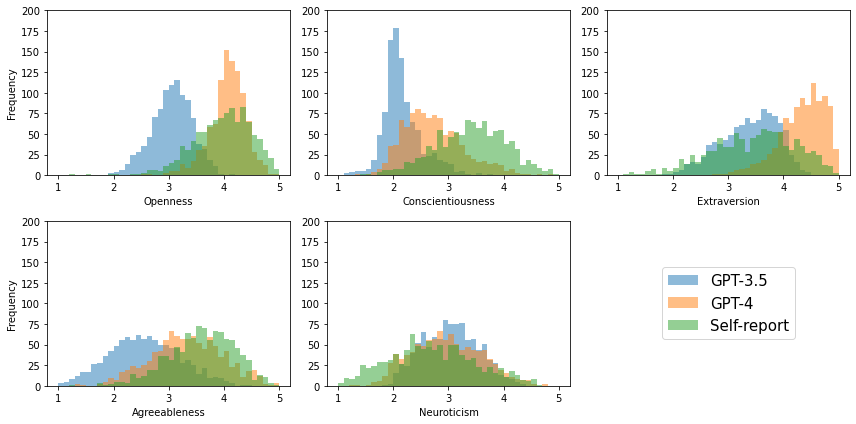}
    \caption{Distributions of self-reported and inferred personality scores for GPT-3.5 and GPT-4. Histograms show absolute frequencies for an overall sample size of n=1000. GPT-3.5 underestimates Openness. Both models underestimate Conscientiousness and Agreeableness but overestimate Neuroticism. For Extraversion, the two models diverge with GPT-3.5 underestimating and GPT-4 overestimating the true scores. Overall, GPT-4 inferred scores were more aligned with self-reported scores, indicating a potential improvement over GPT-3.5.}
    \label{fig:fig_hist}
\end{figure*}

In order to assess the capacity of LLMs to infer psychological traits from social media data, we compared the inferred Big Five personality scores with self-reported scores. A comparison of the distributions suggests that both versions of ChatGPT tended to underestimate Conscientiousness and Agreeableness while overestimating Neuroticism. For Openness and Extraversion, the deviations were inconsistent across ChatGPT versions: While GPT-3.5 tended to underestimate Openness and Extraversion, GPT-4 tended to overestimate Extraversion. Overall, the distributions of inferred scores were more closely aligned with self-reported scores for GPT-4 compared to GPT-3.5, suggesting a potential improvement across versions (see Figure \ref{fig:fig_hist}). Detailed descriptive statistics can be found in S1.

\begin{figure*}[t!]
    \centering
    \includegraphics[width=\textwidth]{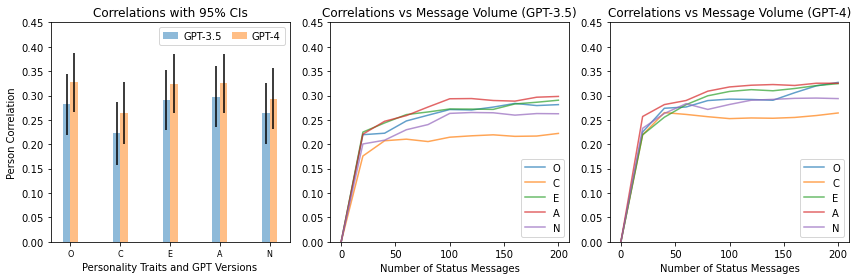}
    \caption{Pearson’s correlation coefficients between inferred and self-reported scores with 95\% confidence intervals (left), and Pearson’s correlation coefficients for GPT-3.5 (mid) and GPT-4 as a function of message volume (right). O: Openness; C: Conscientiousness; E: Extraversion; A: Agreeableness; N: Neuroticism. Inferences for Openness, Extraversion, and Agreeableness were more accurate than those for Conscientiousness and Neuroticism, but the differences remained non-significant. Higher message volume was associated with higher levels of predictive accuracy, but a substantial share of variance was captured in as little as 20 status messages.}
    \label{fig:fig_corall}
\end{figure*}

Importantly, the mere comparison of distributions does not provide insights into the strength and directionality of the relationships between inferred and self-reported scores. For this purpose, we conducted correlation analyses. The average Pearson correlation coefficient of inferred and self-reported scores across all personality traits was r$_{GPT3.5}$=0.27 and r$_{GPT4}$= 0.31. The correlations were highest for the traits of Openness (r$_{GPT3.5}$= .28; r$_{GPT4}$= .33), Extraversion (r$_{GPT3.5}$= .29; r$_{GPT4}$= .32) and and Agreeableness (r$_{GPT3.5}$= .30,  r$_{GPT4}$= .32), and were slightly lower for Conscientiousness (r$_{GPT3.5}$= .22; r$_{GPT4}$= .26) and Neuroticism (r$_{GPT3.5}$= .26; r$_{GPT4}$= .29). All correlation coefficients were significantly different from 0 at p < .001 with Bonferroni correction for multiple comparisons. Similar to the comparison of distributions, GPT-4 showed higher levels of accuracy across all five personality traits, although none of the individual comparisons reached statistical significance (see Figure \ref{fig:fig_corall}). Detailed results, including confidence intervals and significance levels, can be found in S2.

In addition to exploring the capacity of ChatGPT to infer personality traits from social media user data, we also tested the extent to which this capacity is sensitive to changes in the amount of data that was available for inference. Specifically, we computed correlations between self-reported and inferred personality scores based on different numbers of status messages. Specifically, we computed correlations obtained from inferences for a single chunk of status messages (20 status messages) all the way up to ten chunks (200 status messages). As expected, having access to more status messages resulted in more accurate inferences. Notably, however, most correlations are close to their maximum level after observing far less than the ultimate number of 200 status messages. In addition, the inference of certain traits seems to be particularly susceptible to the volume of input data. For example, the models’ accuracy kept increasing with higher levels in input volume for Openness, Extraversion, Agreeableness, and Neuroticism, while the benefits of additional status messages leveled off earlier for Conscientiousness. See Figure \ref{fig:fig_corall} for a graphical representation and S3 for detailed statistics.

\subsection{Does the Quality of LLM Inferences Vary Across Demographic Groups?}
In order to uncover potential gender and age-related biases, we analyzed group differences in inferred Big Five scores, as well as their residuals with respect to self-reported scores. Notably, such gender and age differences might not only emerge in inferred personality scores but are also known to exist in self-reports \cite{feingold_gender_1994, costa_jr_gender_2001}. Consequently, we test for both overall group differences and differences in the residuals between the self-reported and inferred personality scores of each individual.

\subsubsection{Gender Differences}
\begin{figure*}[t]
    \centering
    \includegraphics[width=\textwidth]{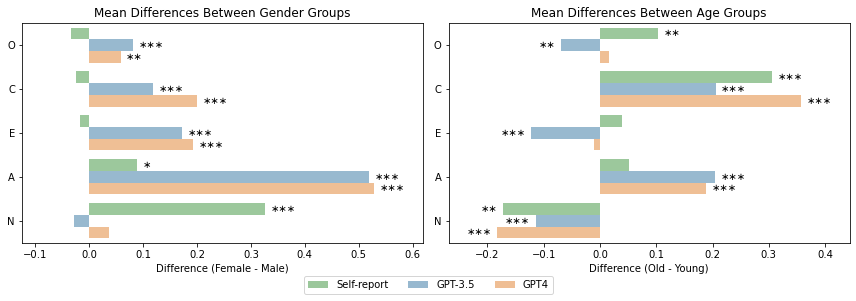}
    \caption{Mean differences in personality scores between gender groups (left) and age groups (right) for self-reported scores as well as inferences by GPT-3.5 and GPT-4. Positive values indicate higher scores for female users compared to male users and older users compared to younger users. O: Openness; C: Conscientiousness; E: Extraversion; A: Agreeableness; N: Neuroticism. ***p<.001; **p<.01; *p<.05. The results show significant gender and age differences across all personality traits.}
    \label{fig:fig_meandiff}
\end{figure*}

We first explored the extent to which any observed group differences in inferred personality traits across men and women aligned with those observed in self-reports. As Figure \ref{fig:fig_meandiff} shows, women tend to score significantly higher in Agreeableness (t=2.31; p=.021) and Neuroticism (t.=6.53; p<.001) when these traits are measured using questionnaires. In contrast, women scored significantly higher in Openness (t$_{GPT3.5}$= 3.42, p$_{GPT3.5}$< .001; t$_{GPT4}$= 2.72, p$_{GPT4}$= .007), Conscientiousness (t$_{GPT3.5}$= 5.28, p$_{GPT3.5}$< .001; t$_{GPT4}$= 5.73, p$_{GPT4}$< .001), Extraversion (t$_{GPT3.5}$= 5.21, p$_{GPT3.5}$< .001; t$_{GPT4}$= 7.25, p$_{GPT4}$< .001), and Agreeableness (t$_{GPT3.5}$= 13.53, p$_{GPT3.5}$< .001; t$_{GPT4}$= 13.63, p$_{GPT4}$< .001) when these traits were inferred by ChatGPT models, with no significant differences found for Neuroticism. This finding offers initial evidence for potential gender biases in the personality inferences made by LLMs (see Figure \ref{fig:fig_meandiff}).

To further explore these potential biases, we analyzed the residuals between inferred scores and self-reported scores as an indication of how well GPT is able to represent the personality traits of male and female users. The findings suggest that GPT’s personality inferences are less accurate for men than women. First, we observed larger absolute residuals for male users in Conscientiousness (t$_{GPT3.5}$= -3.53, p$_{GPT3.5}$< .001; t$_{GPT4}$= -4.48, p$_{GPT4}$< .001), Agreeableness (t$_{GPT3.5}$= -9.22, p$_{GPT3.5}$< .001; t$_{GPT4}$= -5.22, p$_{GPT4}$< .001), and Neuroticism (t$_{GPT3.5}$= -4.55, p$_{GPT3.5}$< .001; t$_{GPT4}$= -2.39, p$_{GPT4}$= .017) across both GPT models, indicating lower accuracy on these traits for men. Additionally, we found larger residuals for male users for GPT-3.5 in Openness (t$_{GPT3.5}$= -3.84, p$_{GPT3.5}$< .001; t$_{GPT4}$= -0.92, p$_{GPT4}$= .357) and larger residuals for female users in Extraversion for GPT-4 (t$_{GPT3.5}$= -1.36, p$_{GPT3.5}$< .173; t$_{GPT4}$= 3.12, p$_{GPT4}$= .002). For a visual representation, please refer to Figure \ref{fig:fig_resdiff}). Detailed statistics can be found in S4.

Taken together, the findings suggest that GPT’s personality inferences are less accurate for men than women. Notably, however, these biases seem to be limited to the absolute measures of accuracy and do not necessarily translate to GPT’s ability to make inferences about men’s relative personality levels. That is, when computing Pearson correlations within gender groups, we did not observe any significant difference in the magnitude of these correlations. Similarly, controlling for gender in the overall correlations between self-reported and inferred personality scores by z-standardizing inferred scores within each gender group did not yield correlations significantly different from those obtained before. 

\subsubsection{Age Differences}
As for gender, we first explored the extent to which any observed group differences in inferred personality traits across younger and older adults (classified using a median split) were aligned with those observed in self-reports. As Figure \ref{fig:fig_meandiff} shows, older users displayed significantly higher self-reported scores in Openness (t=2.96; p=.003) and Conscientiousness (t=7.27; p<.001) and significantly lower self-reported scores in Neuroticism (t=-3.28; p=.001) compared to younger users. Partially mimicking these differences in self-reported personality traits, inferred scores were significantly higher in Conscientiousness (t$_{GPT3.5}$= 9.23, p$_{GPT3.5}$< .001; t$_{GPT4}$= 10.41, p$_{GPT4}$< .001) and Agreeableness (t$_{GPT3.5}$= 4.87, p$_{GPT3.5}$< .001; t$_{GPT4}$= 4.39, p$_{GPT4}$< .001), and lower in Neuroticism (t$_{GPT3.5}$= -3.43, p$_{GPT3.5}$< .001; t$_{GPT4}$= -4.37, p$_{GPT4}$< .001) for older compared to younger users. For Openness (t$_{GPT3.5}$= -2.86, p$_{GPT3.5}$= .004; t$_{GPT4}$= -0.72, p$_{GPT4}$= .472) and Extraversion (t$_{GPT3.5}$= -3.55, p<.001; t$_{GPT4}$= 0.36, p$_{GPT4}$= .717), older individuals scored significantly lower on inferred scores for GPT-3.5 but not GPT-4 (see Figure \ref{fig:fig_meandiff}). 

\begin{figure*}[t]
    \centering
    \includegraphics[width=\textwidth]{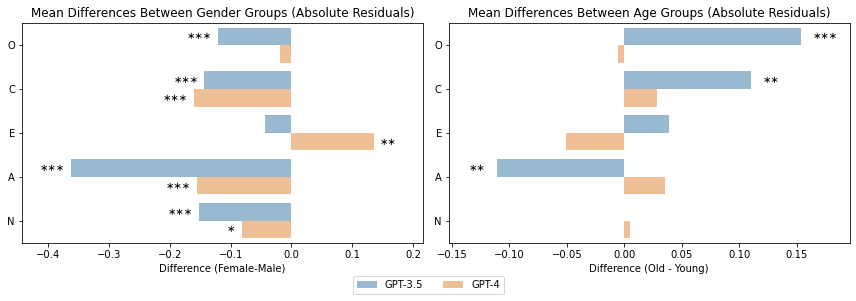}
    \caption{Mean differences in absolute residuals between gender groups (left) and age groups (right) for inferences by GPT-3.5 and GPT-4. Positive values indicate higher residuals for female users compared to male users and older users compared to younger users. O: Openness; C: Conscientiousness; E: Extraversion; A: Agreeableness; N: Neuroticism. ***p<.001; **p<.01; *p<.05. The results indicate lower residuals for female users in all personality traits except Extraversion. Age-related biases were observed for Openness, Conscientiousness, and Agreeableness in inferences by GPT-3.5 but not GPT-4.}
    \label{fig:fig_resdiff}
\end{figure*}

As before, we further explore these differences by analyzing age differences in the residuals between self-reported and inferred scores. Unlike in the analyses of gender, we found substantial inconsistency in the group differences between GPT-3.5 and GPT-4. While the inferences made by GPT-3.5 showed significantly larger absolute residuals for older users in Openness (t$_{GPT3.5}$= 4.78, p$_{GPT3.5}$< .001), Conscientiousness (t$_{GPT3.5}$= 2.64, p$_{GPT3.5}$= .008) and smaller residuals for Agreeableness (t$_{GPT3.5}$= -2.64, p$_{GPT3.5}$= .009), no differences in absolute residuals were found for GPT-4. For a visual representation, please refer to Figure \ref{fig:fig_resdiff}) Detailed statistics can be found in S5.

Taken together, the findings suggest that ChatGPT’s personality inferences might be less accurate for older adults. However, as before, these biases did not translate to ChatGPT’s ability to make inferences about people’s relative personality levels.  We did not find significant differences between within-group correlation coefficients, and z-standardizing personality scores within age groups did not yield correlation coefficients significantly different from those reported before.

\subsection{Agreement With Third-Person Observer Ratings}
We conducted a preliminary analysis examining the correlations between self-reported personality scores and third-person observer ratings, as well as between LLM-inferred scores and third-person observer ratings. This allowed us to 1) compare the quality of LLM inferences against a strong human benchmark (i.e. ratings from people who have access to more identity cues than just social media profiles), and 2) examine the level of agreement between LLM inferences and human judgments. Third-person ratings were collected by letting users' Facebook friends complete a 10-item version of the IPIP personality questionnaire \cite{youyou_computer-based_2015, goldberg_international_2006} about them. The analysis includes a subset of 68 individuals for whom third-person ratings were available. 

The results show that correlations between self-reported scores and observer ratings ranged from r=.198 to r=.378 (mean=.304), while the correlations between LLM-inferred scores and observer ratings ranged from r=.057 to r=.457 (mean=.269) for GPT-3.5 and r=.152 to r=.400 (mean=.276) for GPT-4 (see S6 for detailed results). Overall, the correlation coefficients were largely in the same range as those between self-reported and LLM-inferred scores. The analyses thus suggest that the accuracy of LLM inferences is on par with that of human observers. They also suggest that the LLM is using similar cues to human judges. This is true even though these cues may not always be valid predictors of people’s self-perceptions. For instance, in the case of Conscientiousness, the agreement between LLM inferences and observer ratings was higher than the agreement of either the LLM or human observers with participants’ self-reports.

\section{Discussion}
\subsection{Interpretation of Results}
Our findings suggest that LLMs, such as ChatGPT, can infer psychological dispositions from people’s social media posts without having been explicitly trained to do so. They also offer preliminary evidence that LLMs might generate more accurate inferences for women and younger individuals (compared to men and older adults). Notably, the overall accuracy of the observed inferences (Pearson correlations between self-reported and inferred personality traits ranging between r = .22 and .33, average = .29) is slightly lower than that accomplished by supervised models which have been trained or fine-tuned specifically for this purpose and with the same textual data source as used in testing (e.g., Park et al. \cite{park_automatic_2015}, who reported correlations between r = .26 and r = .41, average r = .37). Yet, the ability of LLMs to produce inferences of reasonably high accuracy in zero-shot learning scenarios has both important theoretical and practical implications.

Our study contributes to a growing body of research comparing the abilities of LLMs to those observed in humans \cite{kosinski_theory_2023, digutsch_overlap_2023, matz_potential_2024}. As our findings suggest, LLMs might have the human-like ability to “profile” people based on their behavioral traces, without ever having had direct interactions with them. Although most social media posts do not contain explicit references to a person’s character, ChatGPT – just like human judges \cite{vazire_e-perceptions_2004, back_facebook_2010} or supervised models \cite{youyou_computer-based_2015} – is able to translate people’s accounts of their daily activities and preferences into a holistic picture of their psychological dispositions. Our results are aligned with previous work suggesting that Openness and Extraversion are more easily inferred than other traits \cite{back_facebook_2010, youyou_computer-based_2015}. At the same time, LLM inferences were more congruent with observer ratings than self-reports in the case of Conscientiousness, indicating that LLMs may also replicate biases in human judgment for certain traits.

Notably, the specific pathways by which LLMs such as ChatGPT arrive at their judgments and the reasons for why certain biases are introduced into the predictions (e.g., systematic gender and age differences) remain unknown. That is, we cannot speak to the question of whether LLMs use the same behavioral cues as humans or supervised machine learning models when translating behavioral residues into psychological profiles or offer an in-depth explanation for the observed differences in accuracy across age and gender categories. For example, the fact that ChatGPT shows systematic biases in its estimation of certain personality traits and is more accurate for women and younger adults could either be indicative of a bias introduced in the training of the models and/or the corpora of text data the models have been trained on, or reflective of differences in people’s general self-expression on social media. 

Specifically, past work indicates that LLMs are susceptible to stereotyping and bias with regard to demographic and geographic groups \cite{bolukbasi_man_2016, abdurahman_perils_2023, durmus_towards_2024, santurkar_whose_2023, atari_which_2023, rathje_gpt_2023}, likely reflecting groups' representation in the underlying training data. At the same time, past work has shown differences in social media use and online self-expression across demographic groups, including age and gender \cite{thayer_online_2006, kondakciu_self-presentation_2021, tifferet_gender_2014, oberst_gender_2016, roberti_female_2022}. While the past literature does not directly speak to differences in personality expression, the observed pattern of results would indicate that women and younger individuals tend to reveal more accurate information about their personalities online.

\subsection{Limitations and Future Research}
Our study has several limitations that should be addressed by future research. Firstly, as mentioned above, the black box character of LLMs prevents us from examining the precise mechanisms by which personality inferences are derived. As a first step in this direction, future research should analyze cue utilization by investigating which language features are highly correlated with inferred trait scores. Similarly, it would be useful to examine which language features are predictive of inference errors in order to better understand the origins of the observed gender and age biases. 

Second, the text data used in our analysis was obtained from the MyPersonality Facebook application \cite{kosinski_facebook_2015}, which was active between 2007 and 2012. Linguistic conventions from this period might differ from contemporary online language, potentially limiting the zero-shot performance of LLMs, which have been trained on newer data. As a result, we would expect the personality inferences of LLMs to be even more accurate when applied to more contemporary data. 

Third, our data was sourced from Facebook users who interacted with the MyPersonality application. As such, our sample might not be representative of the broader population of social media users (or people more generally), which could limit the external validity of our findings. For example, the general underestimation of personality traits such as Openness might be due to the fact that myPersonality users were particularly curious and open-minded. 

Fourth, while our study probed how sensitive the accuracy of LLM-based inferences is to the volume of text input, we limited our data to the 200 most recent status updates. In practice, predictive performance might vary for users with fewer or more status updates. Relatedly, due to the inherent token limit in models like ChatGPT, all input data was processed in chunks. It is possible that the accuracy of future models with the ability to process larger amounts of input data at once might be higher. 

Fifth, our study did not encompass the dynamics of live interactions between LLMs and users. Real-time interactions might yield different insights and highlight additional complexities not captured in our static textual data set \cite{peters_large_2024}. Relatedly, while our research underscores the potential for LLMs in personalizing interactions and enhancing social computing, it does not examine the specifics of how these personalizations can be effectively implemented. 

Sixth, the current research demonstrates the potential of out-of-the-box LLMs for inferring psychological variables using simple techniques such as zero-shot learning and commercially available models. It is likely that the predictive performance of LLMs could be improved through more sophisticated prompting strategies, such as chain-of-thought prompting \cite{yang_psycot_2023} and a combination of in-context learning and supervised fine-tuning \cite{karra_estimating_2023}. While we purposefully focused on zero-shot learning in order to establish a lower bound of predictive accuracy and investigate LLMs' inherent ability to make such predictions, future research could focus on identifying levers that maximize predictive accuracy. Aside from more sophisticated prompting paradigms, this could include giving LLMs access to users’ demographic information which is typically available to human perception and could moderate the interpretation of personality-related signals. For example, the content of status messages may be interpreted differently depending on whether the user is an 18-year-old man or a 55-year-old woman. Being able to interpret message content in the context of sender identity could lead to improved inferences, but could also amplify implicit biases that are known to persist in language models \cite{bolukbasi_man_2016, abdurahman_perils_2023, santurkar_whose_2023}.

Finally, while we make an effort to discuss the societal implications of our findings (see below), detailed recommendations regarding privacy concerns and the potential for misuse should be addressed in future research.

\subsection{Implications}
Our findings also have important practical implications for the application of automated psychological profiling in research and industry. Specifically, the ability of LLMs to infer psychological traits from social media data could foreshadow a remarkable shift in the accessibility - and therefore potential use – of scalable psychometric assessments. For decades, the assessment of psychological traits relied on the use of self-report questionnaires, which are known to be prone to self-report biases and difficult to scale due to their costly and time-consuming nature \cite{podsakoff_sources_2012}. With the introduction of automated psychological assessments driven by supervised machine learning models \cite{kosinski_private_2013, grunenberg_machine_2024}, scientists and practitioners were afforded an alternative approach that promised to expand the study and application of individual differences to research questions and domains that were previously impractical if not impossible (e.g., the use of personality traits in targeted advertising \cite{matz_psychological_2017}; or the investigation of individual differences in large scale, ecologically valid observational studies \cite{freiberg_founder_2023}). However, the widespread application of such automated personality predictions from digital footprints among scientists and practitioners was hindered by the need to collect large amounts of self-report surveys in combination with textual data (see e.g., the myPersonality dataset \cite{kosinski_facebook_2015}) to train and validate the predictive models. With the ability to make similar inferences with models that are available to the broader public, LLMs could democratize access to cheap and scalable psychometric assessments.

While this democratization holds remarkable opportunities for scientific discovery and personalized services, it also introduces considerable ethical challenges. Specifically, the ability to predict people’s intimate psychological needs and preferences without their knowledge or consent poses a threat to people’s privacy and self-determination \cite{matz_privacy_2020}. For instance, users often share information online without considering how this information can be used by third parties and the use of LLMs for psychological profiling may not align with their original intentions. As the case of Cambridge Analytica \cite{hu_cambridge_2020} alongside a growing body of research on personalized persuasion and psychological targeting \cite{matz_psychological_2017, teeny_review_2021, feinberg_moral_2019} has highlighted, insights into people’s psychological dispositions can easily be weaponized to sway opinions and change behavior. Consequently, it might be necessary to introduce guardrails into systems like LLMs that prevent actors from obtaining psychological profiles of thousands or millions of users. Notably, the outlined concerns are aligned with recent calls for regulation \cite{perc_social_2019, hacker_regulating_2023, chan_gpt-3_2023} and the fact that the EU AI Act \cite{european_parliament_artificial_2023} explicitly bans emotion recognition in the workplace and educational institutions, as well as social scoring based on social behavior or personal characteristics.

\subsection{Conclusion}
Taken together, our research demonstrates the capacity of LLMs to derive psychological profiles from social media data, even without specific training. This zero-shot capability underscores the remarkable advancement LLMs represent in the domain of text analysis. While this “intuitive” understanding mirrors distinctly human abilities, the mechanisms and inherent biases associated with LLM-based personality judgments remain elusive and warrant further research. From a practical perspective, the potential of LLMs to effectively infer psychological traits from digital footprints presents a shift in psychometric evaluations, paving the way for large-scale AI-driven assessments. The prospect of democratized, scalable psychometric tools will enable breakthroughs in personalized services and large-scale research. Nevertheless, these advancements bring forth ethical challenges. The potential for non-consensual psychological predictions and other misuses highlights the necessity for stringent ethical frameworks.

\newpage
\section*{Acknowledgments}
We thank the Digital Future Initiative and Columbia Business School for their generous support. We thank Michal Kosinski for fruitful conversations and advice. 

\section*{Author Contributions}
H.P.: Conceptualization, Methodology, Software, Formal analysis, Investigation, Writing - Original Draft, Visualization;
S.C.M.: Conceptualization, Methodology, Writing - Original Draft, Visualization.

\newpage

\printbibliography

@misc{openai_gpt-4_2023,
	title = {{GPT}-4 {Technical} {Report}},
	url = {http://arxiv.org/abs/2303.08774},
	urldate = {2023-08-21},
	publisher = {arXiv},
	author = {OpenAI},
	month = mar,
	year = {2023},
	note = {arXiv:2303.08774 [cs]},
}

@misc{brown_language_2020,
	title = {Language {Models} are {Few}-{Shot} {Learners}},
	url = {http://arxiv.org/abs/2005.14165},
	doi = {10.48550/arXiv.2005.14165},
	urldate = {2023-08-21},
	publisher = {arXiv},
	author = {Brown, Tom B. and Mann, Benjamin and Ryder, Nick and Subbiah, Melanie and Kaplan, Jared and Dhariwal, Prafulla and Neelakantan, Arvind and Shyam, Pranav and Sastry, Girish and Askell, Amanda and Agarwal, Sandhini and Herbert-Voss, Ariel and Krueger, Gretchen and Henighan, Tom and Child, Rewon and Ramesh, Aditya and Ziegler, Daniel M. and Wu, Jeffrey and Winter, Clemens and Hesse, Christopher and Chen, Mark and Sigler, Eric and Litwin, Mateusz and Gray, Scott and Chess, Benjamin and Clark, Jack and Berner, Christopher and McCandlish, Sam and Radford, Alec and Sutskever, Ilya and Amodei, Dario},
	month = jul,
	year = {2020},
	note = {arXiv:2005.14165 [cs]},
}

@inproceedings{radford_language_2019,
	title = {Language {Models} are {Unsupervised} {Multitask} {Learners}},
	url = {https://www.semanticscholar.org/paper/Language-Models-are-Unsupervised-Multitask-Learners-Radford-Wu/9405cc0d6169988371b2755e573cc28650d14dfe},
	urldate = {2023-08-21},
	author = {Radford, Alec and Wu, Jeff and Child, Rewon and Luan, D. and Amodei, Dario and Sutskever, Ilya},
	year = {2019},
}

@inproceedings{bolukbasi_man_2016,
	title = {Man is to {Computer} {Programmer} as {Woman} is to {Homemaker}? {Debiasing} {Word} {Embeddings}},
	volume = {29},
	shorttitle = {Man is to {Computer} {Programmer} as {Woman} is to {Homemaker}?},
	url = {https://papers.nips.cc/paper_files/paper/2016/hash/a486cd07e4ac3d270571622f4f316ec5-Abstract.html},
	urldate = {2023-08-21},
	booktitle = {Advances in {Neural} {Information} {Processing} {Systems}},
	publisher = {Curran Associates, Inc.},
	author = {Bolukbasi, Tolga and Chang, Kai-Wei and Zou, James Y and Saligrama, Venkatesh and Kalai, Adam T},
	year = {2016},
}

@incollection{mccrae_five-factor_2008,
	address = {New York, NY, US},
	title = {The five-factor theory of personality},
	isbn = {978-1-59385-836-0},
	booktitle = {Handbook of personality: {Theory} and research, 3rd ed},
	publisher = {The Guilford Press},
	author = {McCrae, Robert R. and Costa Jr., Paul T.},
	year = {2008},
	pages = {159--181},
}

@article{kosinski_facebook_2015,
	title = {Facebook as a research tool for the social sciences: {Opportunities}, challenges, ethical considerations, and practical guidelines},
	volume = {70},
	issn = {1935-990X},
	shorttitle = {Facebook as a research tool for the social sciences},
	doi = {10.1037/a0039210},
	language = {eng},
	number = {6},
	journal = {The American Psychologist},
	author = {Kosinski, Michal and Matz, Sandra C. and Gosling, Samuel D. and Popov, Vesselin and Stillwell, David},
	month = sep,
	year = {2015},
	pmid = {26348336},
	pages = {543--556},
}

@article{albright_consensus_1988,
	title = {Consensus in personality judgments at zero acquaintance},
	volume = {55},
	issn = {1939-1315},
	doi = {10.1037/0022-3514.55.3.387},
	number = {3},
	journal = {Journal of Personality and Social Psychology},
	author = {Albright, Linda and Kenny, David A. and Malloy, Thomas E.},
	year = {1988},
	note = {Place: US
Publisher: American Psychological Association},
	pages = {387--395},
}

@misc{wan_biasasker_2023,
	title = {{BiasAsker}: {Measuring} the {Bias} in {Conversational} {AI} {System}},
	shorttitle = {{BiasAsker}},
	url = {http://arxiv.org/abs/2305.12434},
	doi = {10.48550/arXiv.2305.12434},
	urldate = {2023-09-05},
	publisher = {arXiv},
	author = {Wan, Yuxuan and Wang, Wenxuan and He, Pinjia and Gu, Jiazhen and Bai, Haonan and Lyu, Michael},
	month = may,
	year = {2023},
	note = {arXiv:2305.12434 [cs]},
}

@article{kosinski_private_2013,
	title = {Private traits and attributes are predictable from digital records of human behavior},
	volume = {110},
	issn = {1091-6490},
	doi = {10.1073/pnas.1218772110},
	number = {15},
	journal = {PNAS Proceedings of the National Academy of Sciences of the United States of America},
	author = {Kosinski, Michal and Stillwell, David and Graepel, Thore},
	year = {2013},
	note = {Place: US
Publisher: National Academy of Sciences},
	pages = {5802--5805},
}

@article{azucar_predicting_2018,
	title = {Predicting the {Big} 5 personality traits from digital footprints on social media: {A} meta-analysis},
	volume = {124},
	issn = {1873-3549},
	shorttitle = {Predicting the {Big} 5 personality traits from digital footprints on social media},
	doi = {10.1016/j.paid.2017.12.018},
	journal = {Personality and Individual Differences},
	author = {Azucar, Danny and Marengo, Davide and Settanni, Michele},
	year = {2018},
	note = {Place: Netherlands
Publisher: Elsevier Science},
	pages = {150--159},
}

@article{schwartz_personality_2013,
	title = {Personality, {Gender}, and {Age} in the {Language} of {Social} {Media}: {The} {Open}-{Vocabulary} {Approach}},
	volume = {8},
	issn = {1932-6203},
	shorttitle = {Personality, {Gender}, and {Age} in the {Language} of {Social} {Media}},
	url = {https://journals.plos.org/plosone/article?id=10.1371/journal.pone.0073791},
	doi = {10.1371/journal.pone.0073791},
	language = {en},
	number = {9},
	urldate = {2023-09-07},
	journal = {PLOS ONE},
	author = {Schwartz, H. Andrew and Eichstaedt, Johannes C. and Kern, Margaret L. and Dziurzynski, Lukasz and Ramones, Stephanie M. and Agrawal, Megha and Shah, Achal and Kosinski, Michal and Stillwell, David and Seligman, Martin E. P. and Ungar, Lyle H.},
	year = {2013},
	note = {Publisher: Public Library of Science},
	pages = {e73791},
}

@article{park_automatic_2015,
	title = {Automatic personality assessment through social media language},
	volume = {108},
	issn = {1939-1315},
	doi = {10.1037/pspp0000020},
	number = {6},
	journal = {Journal of Personality and Social Psychology},
	author = {Park, Gregory and Schwartz, H. Andrew and Eichstaedt, Johannes C. and Kern, Margaret L. and Kosinski, Michal and Stillwell, David J. and Ungar, Lyle H. and Seligman, Martin E. P.},
	year = {2015},
	note = {Place: US
Publisher: American Psychological Association},
	pages = {934--952},
}

@article{yarkoni_personality_2010,
	title = {Personality in 100,000 {Words}: {A} large-scale analysis of personality and word use among bloggers},
	volume = {44},
	issn = {0092-6566},
	shorttitle = {Personality in 100,000 {Words}},
	url = {https://www.ncbi.nlm.nih.gov/pmc/articles/PMC2885844/},
	doi = {10.1016/j.jrp.2010.04.001},
	number = {3},
	urldate = {2023-09-07},
	journal = {Journal of research in personality},
	author = {Yarkoni, Tal},
	month = jun,
	year = {2010},
	pmid = {20563301},
	pmcid = {PMC2885844},
	pages = {363--373},
}

@article{feingold_gender_1994,
	title = {Gender differences in personality: {A} meta-analysis},
	volume = {116},
	issn = {1939-1455},
	shorttitle = {Gender differences in personality},
	doi = {10.1037/0033-2909.116.3.429},
	number = {3},
	journal = {Psychological Bulletin},
	author = {Feingold, Alan},
	year = {1994},
	note = {Place: US
Publisher: American Psychological Association},
	pages = {429--456},
}

@article{costa_jr_gender_2001,
	title = {Gender differences in personality traits across cultures: {Robust} and surprising findings},
	volume = {81},
	issn = {1939-1315},
	shorttitle = {Gender differences in personality traits across cultures},
	doi = {10.1037/0022-3514.81.2.322},
	number = {2},
	journal = {Journal of Personality and Social Psychology},
	author = {Costa Jr., Paul T. and Terracciano, Antonio and McCrae, Robert R.},
	year = {2001},
	note = {Place: US
Publisher: American Psychological Association},
	pages = {322--331},
}

@article{podsakoff_sources_2012,
	title = {Sources of {Method} {Bias} in {Social} {Science} {Research} and {Recommendations} on {How} to {Control} {It}},
	volume = {63},
	url = {https://doi.org/10.1146/annurev-psych-120710-100452},
	doi = {10.1146/annurev-psych-120710-100452},
	number = {1},
	urldate = {2023-09-07},
	journal = {Annual Review of Psychology},
	author = {Podsakoff, Philip M. and MacKenzie, Scott B. and Podsakoff, Nathan P.},
	year = {2012},
	pmid = {21838546},
	note = {\_eprint: https://doi.org/10.1146/annurev-psych-120710-100452},
	pages = {539--569},
}

@article{matz_psychological_2017,
	title = {Psychological targeting as an effective approach to digital mass persuasion},
	volume = {114},
	url = {https://www.pnas.org/doi/10.1073/pnas.1710966114},
	doi = {10.1073/pnas.1710966114},
	number = {48},
	urldate = {2023-09-07},
	journal = {Proceedings of the National Academy of Sciences},
	author = {Matz, S. C. and Kosinski, M. and Nave, G. and Stillwell, D. J.},
	month = nov,
	year = {2017},
	note = {Publisher: Proceedings of the National Academy of Sciences},
	pages = {12714--12719},
}

@article{teeny_review_2021,
	title = {A {Review} and {Conceptual} {Framework} for {Understanding} {Personalized} {Matching} {Effects} in {Persuasion}},
	volume = {31},
	copyright = {© 2020 Society for Consumer Psychology},
	issn = {1532-7663},
	url = {https://onlinelibrary.wiley.com/doi/abs/10.1002/jcpy.1198},
	doi = {10.1002/jcpy.1198},
	language = {en},
	number = {2},
	urldate = {2023-09-07},
	journal = {Journal of Consumer Psychology},
	author = {Teeny, Jacob D. and Siev, Joseph J. and Briñol, Pablo and Petty, Richard E.},
	year = {2021},
	note = {\_eprint: https://onlinelibrary.wiley.com/doi/pdf/10.1002/jcpy.1198},
	pages = {382--414},
}

@article{hu_cambridge_2020,
	title = {Cambridge {Analytica}’s black box},
	volume = {7},
	issn = {2053-9517},
	url = {https://doi.org/10.1177/2053951720938091},
	doi = {10.1177/2053951720938091},
	language = {en},
	number = {2},
	urldate = {2023-09-07},
	journal = {Big Data \& Society},
	author = {Hu, Margaret},
	month = jul,
	year = {2020},
	note = {Publisher: SAGE Publications Ltd},
	pages = {2053951720938091},
}

@misc{hagendorff_thinking_2023,
	title = {Thinking {Fast} and {Slow} in {Large} {Language} {Models}},
	url = {http://arxiv.org/abs/2212.05206},
	doi = {10.48550/arXiv.2212.05206},
	urldate = {2023-09-08},
	publisher = {arXiv},
	author = {Hagendorff, Thilo and Fabi, Sarah and Kosinski, Michal},
	month = aug,
	year = {2023},
	note = {arXiv:2212.05206 [cs]},
}

@article{back_facebook_2010,
	title = {Facebook {Profiles} {Reflect} {Actual} {Personality}, {Not} {Self}-{Idealization}},
	volume = {21},
	doi = {10.1177/0956797609360756},
	journal = {Psychological science},
	author = {Back, Mitja and von der Heiden, Juliane and Vazire, Simine and Gaddis, Sam and Schmukle, Stefan and Egloff, Boris and Gosling, Samuel},
	month = mar,
	year = {2010},
	pages = {372--4},
}

@article{rentfrow_message_2006,
	title = {Message in a {Ballad}: {The} {Role} of {Music} {Preferences} in {Interpersonal} {Perception}},
	volume = {17},
	issn = {0956-7976},
	shorttitle = {Message in a {Ballad}},
	url = {https://doi.org/10.1111/j.1467-9280.2006.01691.x},
	doi = {10.1111/j.1467-9280.2006.01691.x},
	language = {en},
	number = {3},
	urldate = {2023-09-08},
	journal = {Psychological Science},
	author = {Rentfrow, Peter J. and Gosling, Samuel D.},
	month = mar,
	year = {2006},
	note = {Publisher: SAGE Publications Inc},
	pages = {236--242},
}

@article{gosling_room_2002,
	title = {A room with a cue: {Personality} judgments based on offices and bedrooms},
	volume = {82},
	issn = {1939-1315},
	shorttitle = {A room with a cue},
	doi = {10.1037/0022-3514.82.3.379},
	number = {3},
	journal = {Journal of Personality and Social Psychology},
	author = {Gosling, Samuel D. and Ko, Sei Jin and Mannarelli, Thomas and Morris, Margaret E.},
	year = {2002},
	note = {Place: US
Publisher: American Psychological Association},
	pages = {379--398},
}

@misc{kosinski_theory_2023,
	title = {Theory of {Mind} {Might} {Have} {Spontaneously} {Emerged} in {Large} {Language} {Models}},
	url = {http://arxiv.org/abs/2302.02083},
	doi = {10.48550/arXiv.2302.02083},
	urldate = {2023-09-08},
	publisher = {arXiv},
	author = {Kosinski, Michal},
	month = aug,
	year = {2023},
	note = {arXiv:2302.02083 [cs]},
}

@article{digutsch_overlap_2023,
	title = {Overlap in meaning is a stronger predictor of semantic activation in {GPT}-3 than in humans},
	volume = {13},
	copyright = {2023 The Author(s)},
	issn = {2045-2322},
	url = {https://www.nature.com/articles/s41598-023-32248-6},
	doi = {10.1038/s41598-023-32248-6},
	language = {en},
	number = {1},
	urldate = {2023-09-08},
	journal = {Scientific Reports},
	author = {Digutsch, Jan and Kosinski, Michal},
	month = mar,
	year = {2023},
	note = {Number: 1
Publisher: Nature Publishing Group},
	pages = {5035},
}

@article{freiberg_founder_2023,
	title = {Founder personality and entrepreneurial outcomes: {A} large-scale field study of technology startups},
	volume = {120},
	shorttitle = {Founder personality and entrepreneurial outcomes},
	url = {https://www.pnas.org/doi/10.1073/pnas.2215829120},
	doi = {10.1073/pnas.2215829120},
	number = {19},
	urldate = {2023-09-08},
	journal = {Proceedings of the National Academy of Sciences},
	author = {Freiberg, Brandon and Matz, Sandra C.},
	month = may,
	year = {2023},
	note = {Publisher: Proceedings of the National Academy of Sciences},
	pages = {e2215829120},
}

@article{matz_privacy_2020,
	title = {Privacy in the age of psychological targeting},
	volume = {31},
	issn = {2352-250X},
	doi = {10.1016/j.copsyc.2019.08.010},
	journal = {Current Opinion in Psychology},
	author = {Matz, Sandra C. and Appel, Ruth E. and Kosinski, Michal},
	year = {2020},
	note = {Place: Netherlands
Publisher: Elsevier Science},
	pages = {116--121},
}

@article{feinberg_moral_2019,
	title = {Moral reframing: {A} technique for effective and persuasive communication across political divides},
	volume = {13},
	issn = {1751-9004},
	shorttitle = {Moral reframing},
	doi = {10.1111/spc3.12501},
	number = {12},
	journal = {Social and Personality Psychology Compass},
	author = {Feinberg, Matthew and Willer, Robb},
	year = {2019},
	note = {Place: United Kingdom
Publisher: Wiley-Blackwell Publishing Ltd.},
}

@article{goldberg_international_2006,
	title = {The international personality item pool and the future of public-domain personality measures},
	volume = {40},
	issn = {00926566},
	url = {https://linkinghub.elsevier.com/retrieve/pii/S0092656605000553},
	doi = {10.1016/j.jrp.2005.08.007},
	language = {en},
	number = {1},
	urldate = {2023-09-08},
	journal = {Journal of Research in Personality},
	author = {Goldberg, Lewis R. and Johnson, John A. and Eber, Herbert W. and Hogan, Robert and Ashton, Michael C. and Cloninger, C. Robert and Gough, Harrison G.},
	month = feb,
	year = {2006},
	pages = {84--96},
}

@misc{anthropic_model_2023,
	title = {Model {Card} and {Evaluations} for {Claude} {Models}},
	url = {https://www-files.anthropic.com/production/images/Model-Card-Claude-2.pdf},
	publisher = {Anthropic},
	author = {Anthropic},
	year = {2023},
}

@article{peters_large_2024,
	title = {Large {Language} {Models} {Can} {Infer} {Personality} from {Free}-{Form} {User} {Interactions}},
	url = {https://osf.io/apc5g/},
	language = {en-us},
	urldate = {2024-05-19},
	author = {Peters, Heinrich and Cerf, Moran and Matz, Sandra},
	month = may,
	year = {2024},
	note = {Publisher: OSF Preprints},
}

@article{grunenberg_machine_2024,
	title = {Machine learning in recruiting: predicting personality from {CVs} and short text responses},
	volume = {1},
	issn = {2813-7876},
	shorttitle = {Machine learning in recruiting},
	url = {https://www.frontiersin.org/articles/10.3389/frsps.2023.1290295},
	doi = {10.3389/frsps.2023.1290295},
	language = {English},
	urldate = {2024-05-17},
	journal = {Frontiers in Social Psychology},
	author = {Grunenberg, Eric and Peters, Heinrich and Francis, Matt J. and Back, Mitja D. and Matz, Sandra C.},
	month = jan,
	year = {2024},
	note = {Publisher: Frontiers},
}

@article{rathje_gpt_2023,
	title = {{GPT} is an effective tool for multilingual psychological text analysis},
	url = {https://osf.io/sekf5},
	doi = {10.31234/osf.io/sekf5},
	language = {en-us},
	urldate = {2024-05-12},
	journal = {https://osf.io/sekf5},
	author = {Rathje, Steve and Mirea, Dan-Mircea and Sucholutsky, Ilia and Marjieh, Raja and Robertson, Claire and Bavel, Jay J. Van},
	month = may,
	year = {2023},
}

@article{atari_which_2023,
	title = {Which {Humans}?},
	url = {https://osf.io/5b26t},
	doi = {10.31234/osf.io/5b26t},
	language = {en-us},
	urldate = {2024-05-12},
	journal = {https://osf.io/5b26t},
	author = {Atari, Mohammad and Xue, Mona J. and Park, Peter S. and Blasi, Damián and Henrich, Joseph},
	month = sep,
	year = {2023},
}

@article{abdurahman_perils_2023,
	title = {Perils and {Opportunities} in {Using} {Large} {Language} {Models} in {Psychological} {Research}},
	url = {https://osf.io/tg79n},
	doi = {10.31219/osf.io/tg79n},
	language = {en-us},
	urldate = {2024-05-12},
	journal = {https://osf.io/tg79n},
	author = {Abdurahman, Suhaib and Atari, Mohammad and Karimi-Malekabadi, Farzan and Xue, Mona J. and Trager, Jackson and Park, Peter S. and Golazizian, Preni and Omrani, Ali and Dehghani, Morteza},
	month = nov,
	year = {2023},
}

@article{santurkar_whose_2023,
	title = {Whose {Opinions} {Do} {Language} {Models} {Reflect}?},
	url = {http://arxiv.org/abs/2303.17548},
	doi = {10.48550/arXiv.2303.17548},
	urldate = {2024-05-13},
	journal = {arXiv:2303.17548 [cs]},
	author = {Santurkar, Shibani and Durmus, Esin and Ladhak, Faisal and Lee, Cinoo and Liang, Percy and Hashimoto, Tatsunori},
	month = mar,
	year = {2023},
	note = {arXiv:2303.17548 [cs]},
}

@article{durmus_towards_2024,
	title = {Towards {Measuring} the {Representation} of {Subjective} {Global} {Opinions} in {Language} {Models}},
	url = {http://arxiv.org/abs/2306.16388},
	doi = {10.48550/arXiv.2306.16388},
	urldate = {2024-05-13},
	journal = {arXiv:2306.16388 [cs]},
	author = {Durmus, Esin and Nguyen, Karina and Liao, Thomas I. and Schiefer, Nicholas and Askell, Amanda and Bakhtin, Anton and Chen, Carol and Hatfield-Dodds, Zac and Hernandez, Danny and Joseph, Nicholas and Lovitt, Liane and McCandlish, Sam and Sikder, Orowa and Tamkin, Alex and Thamkul, Janel and Kaplan, Jared and Clark, Jack and Ganguli, Deep},
	month = apr,
	year = {2024},
	note = {arXiv:2306.16388 [cs]},
}

@article{karra_estimating_2023,
	title = {Estimating the {Personality} of {White}-{Box} {Language} {Models}},
	url = {http://arxiv.org/abs/2204.12000},
	doi = {10.48550/arXiv.2204.12000},
	urldate = {2024-04-23},
	journal = {arXiv:2204.12000 [cs]},
	author = {Karra, Saketh Reddy and Nguyen, Son The and Tulabandhula, Theja},
	month = may,
	year = {2023},
	note = {arXiv:2204.12000 [cs]},
}

@article{yang_psycot_2023,
	title = {{PsyCoT}: {Psychological} {Questionnaire} as {Powerful} {Chain}-of-{Thought} for {Personality} {Detection}},
	shorttitle = {{PsyCoT}},
	url = {http://arxiv.org/abs/2310.20256},
	doi = {10.48550/arXiv.2310.20256},
	urldate = {2024-04-17},
	journal = {arXiv:2310.20256 [cs]},
	author = {Yang, Tao and Shi, Tianyuan and Wan, Fanqi and Quan, Xiaojun and Wang, Qifan and Wu, Bingzhe and Wu, Jiaxiang},
	month = nov,
	year = {2023},
	note = {arXiv:2310.20256 [cs]},
}

@article{thayer_online_2006,
	title = {Online {Communication} {Preferences} across {Age}, {Gender}, and {Duration} of {Internet} {Use}},
	volume = {9},
	issn = {1094-9313},
	url = {https://www.liebertpub.com/doi/abs/10.1089/cpb.2006.9.432},
	doi = {10.1089/cpb.2006.9.432},
	number = {4},
	urldate = {2024-05-12},
	journal = {CyberPsychology \& Behavior},
	author = {Thayer, Stacy E. and Ray, Sukanya},
	month = aug,
	year = {2006},
	note = {Publisher: Mary Ann Liebert, Inc., publishers},
	pages = {432--440},
}

@article{kondakciu_self-presentation_2021,
	title = {Self-presentation and gender on social media: an exploration of the expression of “authentic selves”},
	volume = {25},
	issn = {1352-2752},
	shorttitle = {Self-presentation and gender on social media},
	url = {https://doi.org/10.1108/QMR-03-2021-0039},
	doi = {10.1108/QMR-03-2021-0039},
	number = {1},
	urldate = {2024-05-12},
	journal = {Qualitative Market Research: An International Journal},
	author = {Kondakciu, Klaudia and Souto, Melissa and Zayer, Linda Tuncay},
	month = jan,
	year = {2021},
	note = {Publisher: Emerald Publishing Limited},
	pages = {80--99},
}

@article{tifferet_gender_2014,
	title = {Gender differences in {Facebook} self-presentation: {An} international randomized study},
	volume = {35},
	issn = {0747-5632},
	shorttitle = {Gender differences in {Facebook} self-presentation},
	url = {https://www.sciencedirect.com/science/article/pii/S0747563214001381},
	doi = {10.1016/j.chb.2014.03.016},
	urldate = {2024-05-12},
	journal = {Computers in Human Behavior},
	author = {Tifferet, Sigal and Vilnai-Yavetz, Iris},
	month = jun,
	year = {2014},
	pages = {388--399},
}

@article{oberst_gender_2016,
	title = {Gender stereotypes in {Facebook} profiles: {Are} women more female online?},
	volume = {60},
	issn = {0747-5632},
	shorttitle = {Gender stereotypes in {Facebook} profiles},
	url = {https://www.sciencedirect.com/science/article/pii/S0747563216301480},
	doi = {10.1016/j.chb.2016.02.085},
	urldate = {2024-05-12},
	journal = {Computers in Human Behavior},
	author = {Oberst, Ursula and Renau, Vanessa and Chamarro, Andrés and Carbonell, Xavier},
	month = jul,
	year = {2016},
	pages = {559--564},
}

@article{roberti_female_2022,
	title = {Female influencers: {Analyzing} the social media representation of female subjectivity in {Italy}},
	volume = {7},
	issn = {2297-7775},
	shorttitle = {Female influencers},
	url = {https://www.frontiersin.org/articles/10.3389/fsoc.2022.1024043},
	doi = {10.3389/fsoc.2022.1024043},
	language = {English},
	urldate = {2024-05-12},
	journal = {Frontiers in Sociology},
	author = {Roberti, Geraldina},
	month = sep,
	year = {2022},
	note = {Publisher: Frontiers},
}

@article{kenny_consensus_1994,
	title = {Consensus in interpersonal perception: {Acquaintance} and the big five},
	volume = {116},
	issn = {1939-1455},
	shorttitle = {Consensus in interpersonal perception},
	doi = {10.1037/0033-2909.116.2.245},
	number = {2},
	journal = {Psychological Bulletin},
	author = {Kenny, David A. and Albright, Linda and Malloy, Thomas E. and Kashy, Deborah A.},
	year = {1994},
	note = {Place: US
Publisher: American Psychological Association},
	pages = {245--258},
}

@article{matz_potential_2024,
	title = {The potential of generative {AI} for personalized persuasion at scale},
	volume = {14},
	copyright = {2024 The Author(s)},
	issn = {2045-2322},
	url = {https://www.nature.com/articles/s41598-024-53755-0},
	doi = {10.1038/s41598-024-53755-0},
	language = {en},
	number = {1},
	urldate = {2024-03-01},
	journal = {Scientific Reports},
	author = {Matz, S. C. and Teeny, J. D. and Vaid, S. S. and Peters, H. and Harari, G. M. and Cerf, M.},
	month = feb,
	year = {2024},
	note = {Publisher: Nature Publishing Group},
	pages = {4692},
}

@article{vazire_e-perceptions_2004,
	title = {e-{Perceptions}: {Personality} {Impressions} {Based} on {Personal} {Websites}},
	volume = {87},
	issn = {1939-1315},
	shorttitle = {e-{Perceptions}},
	doi = {10.1037/0022-3514.87.1.123},
	number = {1},
	journal = {Journal of Personality and Social Psychology},
	author = {Vazire, Simine and Gosling, Samuel D.},
	year = {2004},
	pages = {123--132},
}

@inproceedings{hacker_regulating_2023,
	title = {Regulating {ChatGPT} and other {Large} {Generative} {AI} {Models}},
	isbn = {9798400701924},
	url = {https://dl.acm.org/doi/10.1145/3593013.3594067},
	doi = {10.1145/3593013.3594067},
	urldate = {2024-01-24},
	booktitle = {Proceedings of the 2023 {ACM} {Conference} on {Fairness}, {Accountability}, and {Transparency}},
	author = {Hacker, Philipp and Engel, Andreas and Mauer, Marco},
	year = {2023},
	pages = {1112--1123},
}

@article{perc_social_2019,
	title = {Social and juristic challenges of artificial intelligence},
	volume = {5},
	issn = {2055-1045},
	url = {https://www.nature.com/articles/s41599-019-0278-x},
	doi = {10.1057/s41599-019-0278-x},
	language = {en},
	number = {1},
	urldate = {2024-01-24},
	journal = {Palgrave Communications},
	author = {Perc, Matjaž and Ozer, Mahmut and Hojnik, Janja},
	year = {2019},
	pages = {s41599--019--0278--x},
}

@misc{european_parliament_artificial_2023,
	title = {Artificial {Intelligence} {Act}: deal on comprehensive rules for trustworthy {AI}},
	shorttitle = {Artificial {Intelligence} {Act}},
	url = {https://www.europarl.europa.eu/news/en/press-room/20231206IPR15699/artificial-intelligence-act-deal-on-comprehensive-rules-for-trustworthy-ai},
	language = {en},
	urldate = {2024-01-25},
	author = {European Parliament},
	year = {2023},
}

@article{chan_gpt-3_2023,
	title = {{GPT}-3 and {InstructGPT}: technological dystopianism, utopianism, and “{Contextual}” perspectives in {AI} ethics and industry},
	volume = {3},
	issn = {2730-5961},
	shorttitle = {{GPT}-3 and {InstructGPT}},
	url = {https://doi.org/10.1007/s43681-022-00148-6},
	doi = {10.1007/s43681-022-00148-6},
	language = {en},
	number = {1},
	urldate = {2024-01-24},
	journal = {AI and Ethics},
	author = {Chan, Anastasia},
	year = {2023},
	pages = {53--64},
}

@article{youyou_computer-based_2015,
	title = {Computer-based personality judgments are more accurate than those made by humans},
	volume = {112},
	issn = {0027-8424},
	url = {https://www.ncbi.nlm.nih.gov/pmc/articles/PMC4313801/},
	doi = {10.1073/pnas.1418680112},
	number = {4},
	urldate = {2022-01-25},
	journal = {Proceedings of the National Academy of Sciences of the United States of America},
	author = {Youyou, Wu and Kosinski, Michal and Stillwell, David},
	month = jan,
	year = {2015},
	pmid = {25583507},
	pmcid = {PMC4313801},
	pages = {1036--1040},
}

\end{document}


\maketitle

\appendix

\section{Descriptive Statistics}
\label{app:table_descriptives}

\begin{tabular*}{\textwidth}{@{\extracolsep{\fill}} l l r r r r r r r r}
\toprule
            &   &   count &   mean &    std &    min &    25\% &    50\% &    75\% &    max \\
score & trait &         &        &        &        &        &        &        &        \\
\midrule
GPT-3.5 & O &  1000.0 &  3.074 &  0.364 &  1.933 &  2.833 &  3.100 &  3.333 &  4.133 \\
            & C &  1000.0 &  2.177 &  0.347 &  1.133 &  1.967 &  2.100 &  2.333 &  3.800 \\
            & E &  1000.0 &  3.401 &  0.511 &  1.833 &  3.058 &  3.467 &  3.800 &  4.500 \\
            & A &  1000.0 &  2.581 &  0.637 &  1.000 &  2.133 &  2.567 &  3.033 &  4.500 \\
            & N &  1000.0 &  3.056 &  0.499 &  1.900 &  2.667 &  3.067 &  3.408 &  4.400 \\
\midrule[0.8pt]
GPT-4 & O &  1000.0 &  4.085 &  0.326 &  2.533 &  3.933 &  4.100 &  4.300 &  4.967 \\
            & C &  1000.0 &  2.741 &  0.543 &  1.333 &  2.367 &  2.667 &  3.067 &  4.833 \\
            & E &  1000.0 &  4.354 &  0.415 &  2.800 &  4.100 &  4.400 &  4.667 &  5.000 \\
            & A &  1000.0 &  3.345 &  0.644 &  1.333 &  2.933 &  3.333 &  3.800 &  4.967 \\
            & N &  1000.0 &  2.979 &  0.625 &  1.300 &  2.500 &  2.950 &  3.433 &  4.800 \\
\midrule[0.8pt]
Self-report & O &  1000.0 &  3.998 &  0.518 &  1.300 &  3.650 &  4.050 &  4.400 &  5.000 \\
            & C &  1000.0 &  3.431 &  0.640 &  1.500 &  3.000 &  3.450 &  3.900 &  5.000 \\
            & E &  1000.0 &  3.473 &  0.766 &  1.100 &  2.950 &  3.550 &  4.050 &  5.000 \\
            & A &  1000.0 &  3.613 &  0.589 &  1.300 &  3.250 &  3.650 &  4.050 &  5.000 \\
            & N &  1000.0 &  2.742 &  0.779 &  1.000 &  2.200 &  2.700 &  3.250 &  4.700 \\
\bottomrule
\end{tabular*}
\vspace{5pt}

Descriptive statistics for Big Five personality scores by GPT-3.5, GPT-4, and self-report (IPIP). O: Openness; C: Conscientiousness; E: Extraversion; A: Agreeableness; N: Neuroticism.

\vspace{10pt}

\section{Correlation Analyses}
\label{app:table_corr}

\begin{tabular*}{\textwidth}{@{\extracolsep{\fill}}lrrrrlr}
\toprule
{} &    cor &   ci\_l &   ci\_u &    p &    pers &    err \\
version &        &        &        &      &         &        \\
\midrule
GPT-3.5 &  0.282 &  0.217 &  0.344 &  0.0 &  o\_pers &  0.062 \\
GPT-3.5 &  0.223 &  0.156 &  0.287 &  0.0 &  c\_pers &  0.065 \\
GPT-3.5 &  0.291 &  0.226 &  0.353 &  0.0 &  e\_pers &  0.062 \\
GPT-3.5 &  0.298 &  0.234 &  0.360 &  0.0 &  a\_pers &  0.062 \\
GPT-3.5 &  0.263 &  0.198 &  0.326 &  0.0 &  n\_pers &  0.063 \\
\midrule[0.8pt]
GPT-4   &  0.327 &  0.264 &  0.387 &  0.0 &  o\_pers &  0.060 \\
GPT-4   &  0.264 &  0.199 &  0.327 &  0.0 &  c\_pers &  0.063 \\
GPT-4   &  0.324 &  0.261 &  0.385 &  0.0 &  e\_pers &  0.060 \\
GPT-4   &  0.325 &  0.262 &  0.385 &  0.0 &  a\_pers &  0.060 \\
GPT-4   &  0.294 &  0.229 &  0.355 &  0.0 &  n\_pers &  0.062 \\
\bottomrule
\end{tabular*}
\vspace{5pt}

Correlations between inferred and self-reported personality scores for GPT-3.5 and GPT-4.

\section{Correlations as a Function of Input Volume}
\label{app:corhist}
\begin{tabular*}{\textwidth}{@{\extracolsep{\fill}} lrrrrrr}
\toprule
chunk &   messages &      O &      C &      E &      A &      N \\
\midrule
0  &    0.0 &  0.000 &  0.000 &  0.000 &  0.000 &  0.000 \\
1  &   20.0 &  0.220 &  0.176 &  0.225 &  0.221 &  0.201 \\
2  &   40.0 &  0.223 &  0.207 &  0.244 &  0.247 &  0.208 \\
3  &   60.0 &  0.248 &  0.211 &  0.261 &  0.259 &  0.230 \\
4  &   80.0 &  0.260 &  0.206 &  0.266 &  0.276 &  0.240 \\
5  &  100.0 &  0.271 &  0.215 &  0.272 &  0.293 &  0.263 \\
6  &  120.0 &  0.270 &  0.217 &  0.272 &  0.294 &  0.265 \\
7  &  140.0 &  0.276 &  0.219 &  0.272 &  0.290 &  0.265 \\
8  &  160.0 &  0.284 &  0.216 &  0.283 &  0.289 &  0.260 \\
9  &  180.0 &  0.279 &  0.217 &  0.286 &  0.297 &  0.263 \\
10 &  200.0 &  0.281 &  0.222 &  0.290 &  0.298 &  0.263 \\
\bottomrule
\end{tabular*}
\vspace{10pt}

Correlations as a function of input volume for inferences by GPT-3.5. O: Openness; C: Conscientiousness; E: Extraversion; A: Agreeableness; N: Neuroticism.

\vspace{20pt}

\begin{tabular*}{\textwidth}{@{\extracolsep{\fill}} lrrrrrr}
\toprule
chunk &   messages &      O &      C &      E &      A &      N \\
\midrule
0  &    0.0 &  0.000 &  0.000 &  0.000 &  0.000 &  0.000 \\
1  &   20.0 &  0.225 &  0.219 &  0.219 &  0.257 &  0.233 \\
2  &   40.0 &  0.274 &  0.265 &  0.256 &  0.281 &  0.263 \\
3  &   60.0 &  0.276 &  0.261 &  0.281 &  0.290 &  0.283 \\
4  &   80.0 &  0.290 &  0.257 &  0.299 &  0.309 &  0.272 \\
5  &  100.0 &  0.293 &  0.253 &  0.308 &  0.318 &  0.282 \\
6  &  120.0 &  0.292 &  0.254 &  0.312 &  0.321 &  0.290 \\
7  &  140.0 &  0.290 &  0.253 &  0.310 &  0.323 &  0.292 \\
8  &  160.0 &  0.306 &  0.255 &  0.314 &  0.321 &  0.294 \\
9  &  180.0 &  0.320 &  0.259 &  0.320 &  0.325 &  0.295 \\
10 &  200.0 &  0.327 &  0.264 &  0.324 &  0.325 &  0.294 \\
\bottomrule
\end{tabular*}
\vspace{10pt}

Correlations as a function of input volume for inferences by GPT-4. O: Openness; C: Conscientiousness; E: Extraversion; A: Agreeableness; N: Neuroticism.
\newpage

\section{Subgroup Analysis - Gender}
\label{app:table_gender}

\begin{tabular*}{\textwidth}{@{\extracolsep{\fill}} llrrrrrr}
\toprule
& & \multicolumn{3}{c}{GPT-3.5} & \multicolumn{3}{c}{GPT-4} \\
\cmidrule(lr){3-5} \cmidrule(lr){6-8}
& & d & t & p & d & t & p \\
\midrule[0.8pt]
\textbf{Self-report scores}
& O & -0.067 &  -1.022 & -0.307 & -0.067 &  -1.022 & -0.307 \\
& C & -0.038 &  -0.578 & -0.563 & -0.038 &  -0.578 & -0.563 \\
& E & -0.022 &  -0.336 & -0.737 & -0.022 &  -0.336 & -0.737 \\
& A &  0.151 &   2.309 & -0.021 &  0.151 &   2.309 & -0.021 \\
& N &  0.428 &   6.530 & -0.000 &  0.428 &   6.530 & -0.000 \\
\midrule[0.8pt]
\textbf{Inferred scores}
& O &  0.224 &   3.419 & -0.001 &  0.179 &   2.724 & -0.007 \\
& C &  0.346 &   5.280 & -0.000 &  0.375 &   5.729 & -0.000 \\
& E &  0.342 &   5.214 & -0.000 &  0.475 &   7.250 & -0.000 \\
& A &  0.887 &  13.530 & -0.000 &  0.894 &  13.634 & -0.000 \\
& N & -0.056 &  -0.861 & -0.389 &  0.060 &   0.917 & -0.360 \\
\midrule[0.8pt]
\textbf{Absolute residuals}
& O & -0.251 &  -3.827 & -0.000 & -0.060 &  -0.921 & -0.357 \\
& C & -0.232 &  -3.534 & -0.000 & -0.294 &  -4.481 & -0.000 \\
& E & -0.089 &  -1.363 & -0.173 &  0.208 &   3.181 & -0.002 \\
& A & -0.605 &  -9.224 & -0.000 & -0.342 &  -5.218 & -0.000 \\
& N & -0.298 &  -4.553 & -0.000 & -0.157 &  -2.394 & -0.017 \\
\midrule[0.8pt]
\textbf{Directed residuals}
& O &  0.214 &   3.270 & -0.001 &  0.181 &   2.762 & -0.006 \\
& C &  0.219 &   3.335 & -0.001 &  0.315 &   4.811 & -0.000 \\
& E &  0.242 &   3.690 & -0.000 &  0.284 &   4.331 & -0.000 \\
& A &  0.618 &   9.428 & -0.000 &  0.641 &   9.786 & -0.000 \\
& N & -0.450 &  -6.864 & -0.000 & -0.347 &  -5.300 & -0.000 \\
\bottomrule
\end{tabular*}
\vspace{10pt}

Comparisons of self-report scores, inferred scores, absolute residuals, and directed residuals across gender groups. A positive test statistic indicates a higher group mean for female users. O: Openness; C: Conscientiousness; E: Extraversion; A: Agreeableness; N: Neuroticism.
\newpage

\section{Subgroup Analysis - Age}
\label{app:table_age}

\begin{tabular*}{\textwidth}{@{\extracolsep{\fill}} llrrrrrr}
\toprule
& & \multicolumn{3}{c}{GPT-3.5} & \multicolumn{3}{c}{GPT-4} \\
\cmidrule(lr){3-5} \cmidrule(lr){6-8}
& & d & t & p & d & t & p \\
\midrule[0.8pt]
\textbf{Self-report scores}
& O &  0.198 &  2.961 & -0.003 &  0.198 &   2.961 & -0.003 \\
& C &  0.487 &  7.276 & -0.000 &  0.487 &   7.276 & -0.000 \\
& E &  0.050 &  0.751 & -0.453 &  0.050 &   0.751 & -0.453 \\
& A &  0.088 &  1.312 & -0.190 &  0.088 &   1.312 & -0.190 \\
& N & -0.220 & -3.283 & -0.001 & -0.220 &  -3.283 & -0.001 \\
\midrule[0.8pt]
\textbf{Inferred scores}
& O & -0.192 & -2.860 & -0.004 &  0.048 &   0.720 & -0.472 \\
& C &  0.618 &  9.228 & -0.000 &  0.697 &  10.414 & -0.000 \\
& E & -0.237 & -3.546 & -0.000 & -0.024 &  -0.363 & -0.717 \\
& A &  0.326 &  4.867 & -0.000 &  0.294 &   4.388 & -0.000 \\
& N & -0.230 & -3.429 & -0.001 & -0.292 &  -4.368 & -0.000 \\
\midrule[0.8pt]
\textbf{Absolute residuals}
& O &  0.320 &  4.777 & -0.000 & -0.017 &  -0.253 & -0.801 \\
& C &  0.177 &  2.638 & -0.008 &  0.052 &   0.769 & -0.442 \\
& E &  0.080 &  1.201 & -0.230 & -0.077 &  -1.150 & -0.251 \\
& A & -0.177 & -2.636 & -0.009 &  0.076 &   1.140 & -0.255 \\
& N & -0.001 & -0.010 & -0.992 &  0.009 &   0.132 & -0.895 \\
\midrule[0.8pt]
\textbf{Directed residuals}
& O & -0.318 & -4.750 & -0.000 & -0.171 &  -2.548 & -0.011 \\
& C & -0.152 & -2.266 & -0.024 &  0.071 &   1.064 & -0.288 \\
& E & -0.203 & -3.031 & -0.003 & -0.066 &  -0.979 & -0.328 \\
& A &  0.212 &  3.161 & -0.002 &  0.192 &   2.860 & -0.004 \\
& N &  0.072 &  1.080 & -0.280 & -0.011 &  -0.169 & -0.866 \\
\bottomrule
\end{tabular*} 
\vspace{10pt}

Comparisons of self-report scores, inferred scores, absolute residuals, and directed residuals across age groups. A positive test statistic indicates a higher group mean for users of above-median age. O: Openness; C: Conscientiousness; E: Extraversion; A: Agreeableness; N: Neuroticism.